%% file: Version_6_RSS.tex
\documentclass[conference]{IEEEtran}

\input{Preamble.tex}

\usepackage[numbers]{natbib}

\input{title_authors.tex}

\begin{document}
\maketitle
\begin{abstract}
Registration of 3D point clouds is a fundamental task in several applications of robotics and computer vision. While registration methods such as iterative closest point and variants are very popular, they are only locally optimal. There has been some recent work on globally optimal registration, but they perform poorly in the presence of noise in the measurements. In this work we develop a mixed integer programming-based approach for globally optimal registration that explicitly considers uncertainty in its optimization, and hence produces more accurate estimates. Furthermore, from a practical implementation perspective we develop a multi-step optimization that combines fast local methods with our accurate global formulation. Through extensive simulation and real world experiments we demonstrate improved performance over state-of-the-art methods for various level of noise and outliers in the data as well as for partial geometric overlap.
\end{abstract}

\input{introduction.tex}
\input{relatedwork.tex}
\input{notation.tex}
\input{modeling2.tex}

\input{results.tex}

\input{conclusion.tex}

\appendix
\input{appendix.tex}


\bibliography{references}
\bibliographystyle{plainnat}

\end{document}

%% file: Preamble.tex
\usepackage{amsmath,mathtools} 
\usepackage{amssymb}  
\usepackage{graphicx} 
\usepackage{cite} 
\usepackage{authblk}
\usepackage{multirow}
\usepackage[linesnumbered,ruled,vlined]	{algorithm2e}
\usepackage{algpseudocode}
\usepackage{booktabs}
\usepackage{enumerate}
\usepackage{color}
\usepackage{bbm}
\usepackage{dsfont}
\usepackage{hyperref}

\newcommand{\realfield}{\hbox{I \kern -.35em R}}

\newcommand{\bs}[1]{\ensuremath{\boldsymbol{#1}}}

\newcommand{\bt}{\bs{t}\xspace}

\newcommand{\bo}{\bs{o}\xspace}

\newcommand{\bv}{\bs{v}\xspace}

\newcommand{\bss}{\bs{s}\xspace}
\newcommand{\mM}{\mathcal{M}\xspace}

\newcommand{\mV}{\mathcal{V}\xspace}
\newcommand{\mS}{\mathcal{S}\xspace}

\newcommand{\bI}{\bs{I}\xspace}

\newcommand{\bC}{\bs{C}\xspace}

\newcommand{\bm}{\bs{m}\xspace}

\newcommand{\bB}{\bs{B}\xspace}
\newcommand{\bu}{\bs{u}\xspace}

\newcommand{\bQ}{\bs{Q}\xspace}

\newcommand{\etal}{\emph{et al.}\xspace}

\newcommand{\bH}{\bs{H}\xspace}
\newcommand{\bSigma}{\bs{\Sigma}\xspace}
\newcommand{\Real}{\mathbb{R}\xspace}
\newcommand{\bR}{\bs{R}\xspace}

\newcommand{\SO}{\ensuremath{SO(3)}\xspace}

\newcommand{\NsSampled}{N_{s}}

\newcommand{\corrFive}{\bH}
\newcommand{\corrFivePntTwo}{\bH'}
\newcommand{\phimaxCorr}{\phi_{max1}}
\newcommand{\phimaxOut}{\phi_{max2}}
\newcommand{\covariance}{\bC}
\newcommand{\chol}{\bB}
\newcommand{\TejasFive}{\mbox{PCR-MIP-Mah}\xspace}

\newcommand{\TejasOne}{\mbox{PCR-MIP-Eu}\xspace}
\newcommand{\APE}{\mbox{APE}\xspace}
\newcommand{\RN}{\mbox{RN}\xspace}
\newcommand{\LDR}{\mbox{LDR}\xspace}
\newcommand{\modelTolBandLength}{N_{b}}

%% file: title_authors.tex
\title{\LARGE \bf Globally optimal registration of noisy point clouds}

\author[2]{Rangaprasad Arun Srivatsan}
\author[1]{Tejas Zodage}
\author[1]{Howie Choset}
\affil[1]{Robotics Institute, Carnegie Mellon University}
\affil[2]{Apple Inc.}
\affil[ ]{email: aruns@apple.com}

%% file: introduction.tex
\section{Introduction}
\label{sc:intro}
Point cloud registration (PCR) is the problem of finding the transformation that would align point clouds obtained in different frames. This problem is of great importance to the computer vision and robotics community, for instance, estimating pose of objects from lidar measurements~\cite{ding2008automatic}, performing 3D reconstruction~\cite{newcombe2011kinectfusion}, simultaneous localization and mapping~\cite{holz2010sancta}, robot grasping and manipulation~\cite{xiang2017posecnn}, etc. While there exist several methods to perform registration, most notably iterative closest point (ICP) and its variants~\cite{ICP,Szymon01}, they are not robust to large initial misalignment, noise and outliers. 

Using additional information, such as color/intensity~\cite{Godin94}, or feature descriptors~\cite{gelfand2005robust,rusu2009fast}, can help obtain globally optimal registration. But additional information may not be always available (for instance color information is not available when using lidar) and hence we restrict ourselves to applications where only point cloud information is available. Optimizers such as genetic algorithms~\cite{seixas2008image}, simulated annealing~\cite{luck2000registration}, multiple initial start techniques~\cite{srivatsan2016multiple}, branch and bound techniques~\cite{Yang13}, etc. have been used for registration, with weak optimality guarantees~\cite{izattglobally}. An important reason for the weak guaranty is the lack of explicit optimization over point correspondences.

Recently, Izatt~\etal~\cite{izattglobally} developed a method for global registration using mixed integer programming (MIP). This is a unique approach in that it explicitly reasons about optimal registration parameters as well as correspondence variables. As a result they demonstrate improved results over other global registration approaches especially in the presence of outliers. Their approach to registration uses MIP to minimize a Euclidean distance error between the points~\footnote{We shall refer to this method as point cloud registration using mixed integer programming with Euclidean distance function (\TejasOne)}. Two major drawbacks of this approach are -- (1) lack of robustness to noise, and (2) computationally expensive to scale with increase in number of points. 

\begin{figure}[t!]
 	\centering
 	\includegraphics [width=0.5\textwidth]
 	{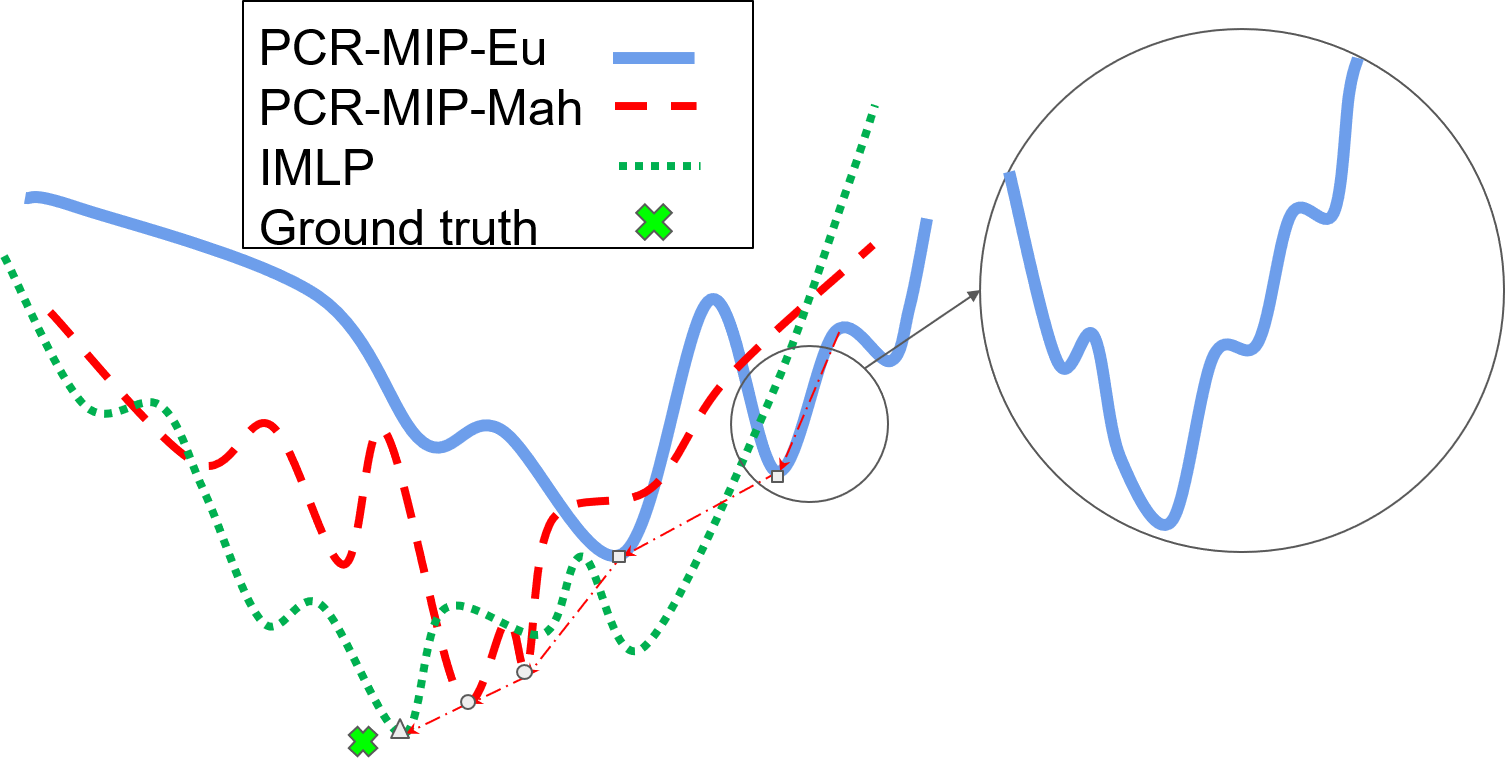}
 	\caption{An iconic illustration of optimization functions used for registration by different methods. First we find an approximate pose using \TejasOne~\cite{izattglobally}, as shown by the blue curve. Following this, we refine the estimate which reasons for uncertainties using our approach, \TejasFive, as shown by the red dashed curve. Finally, we perform a local refinement using IMLP~\cite{Seth15} as shown by the green dotted curve. Note than using a multi-step approach allows us to get to a solution very close to the ground truth.  }
 	\label{fg:optimal}
 \end{figure}

In this work, we develop a registration approach that can handle noise, outliers and partial data, while ensuring global optimality (see Fig.~\ref{fg:optimal}). We improve \TejasOne by incorporating sensor and model uncertainty in the optimization, thus making it robust to noise. To the best of our knowledge, this is one of the first works to provide globally optimal solution in the presence of noise uncertainties. Another improvement we offer is a multi-step optimization process to improve computational time, while allowing for registration using greater number of points compared to \TejasOne.
 
 By conducting extensive tests on both synthetic and real-world data, we demonstrate superior performance over other global registration methods in terms of accuracy as well as robustness. The remainder of the paper is organized as follows -- Section~\ref{sc:related_work} describes the various related works, Sec.~\ref{sc:modeling} describes our formulation. The results are presented in Sec.~\ref{sc:results}, and conclusions are presented in Sec.~\ref{sc:conclusion}.
 
%% file: relatedwork.tex
\section{Related Work}
\label{sc:related_work}
 Besl~\etal~\cite{ICP} introduced the ICP, which is widely used for registration applications in robotics and computer vision. ICP iteratively estimates the transformation by alternating between, (1) estimating the closest-point correspondences  given the current transformation, and (2) estimating the transformation using the current correspondences, until convergence.  Several variants of the ICP have been developed~(see \cite{Szymon01} for a review). Some of the variants incorporate sensor uncertainties~\cite{estepar2004robust,segal2009generalized}, are robust to outliers, incorporate uncertainty in finding correspondence~\cite{Seth15}, are robust to outliers~\cite{tsin2004correlation,phillips2007outlier}, etc. These methods however, need a good initial alignment for convergence.

Using heuristics to find correspondence can improve the robustness of ICP to initial misalignment. However, heuristics are application dependent and what works for one application may not for another. For instance, in probing-based registration for surgical applications, anatomical segments and features can be easily identified by visual inspection. Probing at locations within these anatomical segments can greatly help improve the correspondence as shown in~\cite{simon1995techniques, ma2003robust}. The works of Gelfand~\etal~\cite{gelfand2005robust} use scale invariant curvature features, Glover~\etal~\cite{glover2012} use oriented features and Bingham Procrustean alignment, Makadia~\etal~\cite{makadia2006fully} use extended Gaussian images, Rusu~\etal~\cite{rusu2009fast} use fast point feature histograms, and Godin~\etal~\cite{Godin94} use color intensity information for correspondence matching. When dealing with applications where volumetric data is available, curve-skeletons~\cite{cornea20053d} and heat kernel signature~\cite{ovsjanikov2010one} can be used to obtain a good initial estimate for the pose. 

With the advent for large volumes of easily shareable labeled-datasets, learning-based approaches have gained popularity recently. Learning-based approaches provide good initial pose estimates and rely on local optimizers such an ICP for refined estimates~\cite{tang2013real, vongkulbhisal2017discriminative,kendall2015posenet, xiang2016objectnet3d}. However, these methods generalize poorly to unseen object instances~\cite{balntas2017pose}.  

In order to widen the basin of convergence, Fitzgibbon~\etal~\cite{fitzgibbon2003robust} developed a Levenberg-Marquardt-based approach. Genetic algorithms and simulated annealing have been used in tandem with ICP to help escape local minima~\cite{seixas2008image,luck2000registration}. RANSAC-based hypothesis testing approaches have also been developed, such as the 4PCS~\cite{aiger20084}. Yang~\etal~\cite{Yang13} introduced GoICP, a branch and bound-based optimization approach to obtain globally optimal registration. More recently convex relaxation has been used for global pose estimation using Riemannian optimization~\cite{rosen2016certifiably}, semi-definite programming~\cite{horowitz2014convex,maron2016point} and mixed integer programming~\cite{izattglobally}. A major drawback of the above methods is that none of them consider uncertainty in the points, and are hence not robust to noisy measurements. 

%% file: notation.tex
\begin{table}[htbp]
\caption{Notations}
\label{tb:notations}
\centering
\begin{tabular}{ c l }
\toprule
Symbol&Description\\ \midrule
$N_{\mS}$& number of sensor points\\
$N_{\mV}$& number of model vertices\\
$N_{\mM}$& number of model points\\
$\bss_i \in \Real^3$& Sensor point\\
$\bSigma_{\bss_i}\in \Real^{3 \times 3}$& Uncertainty associated with $\bss_i$\\
$\mV\in \Real^{3\times N_{\mV}}$&Set of vertices of a triangular mesh model\\
$\bC\in \Real^{ N_{\mS} \times N_{\mV}}$&sensor points to model vertices assignment \\
$\mM\in \Real^{ 3 \times N_{\mM}}$&Points generated on the mesh model\\
$\bC^b\in \Real^{ N_{\mS} \times N_{\mM}}$&sensor points to model points assignment\\
$\bo\in \Real^{ N_{\mS}}$&Indicates if a sensor point is an outlier\\\bottomrule
\end{tabular}
\end{table}

%% file: modeling2.tex
\section{Modeling}
\label{sc:modeling}

\begin{figure*}[t!]
	\centering
	\includegraphics [width=1\textwidth]
	{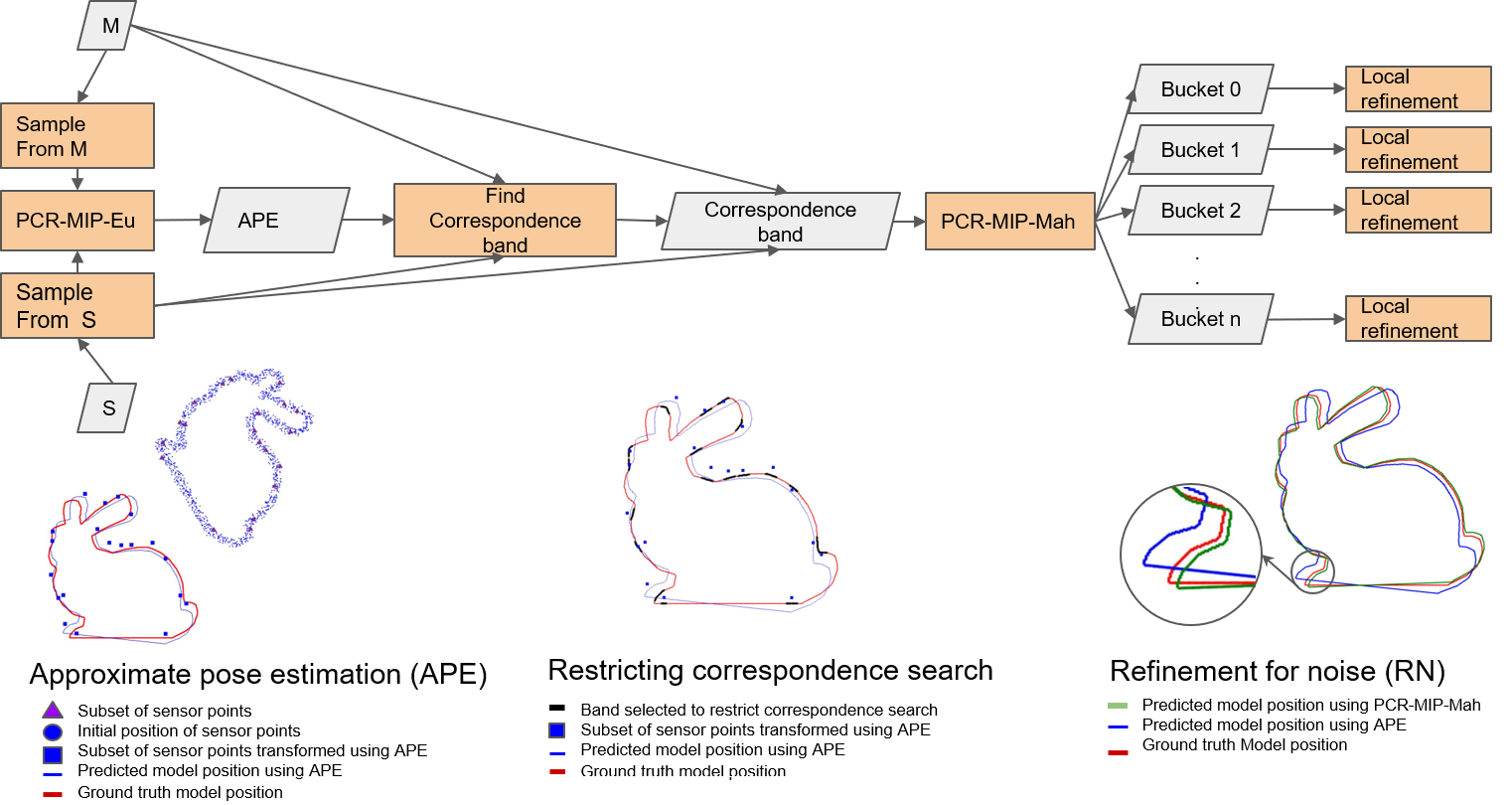}
	\caption{The flowcharts shows various components of the multi-step optimization process that we follow to obtain a globally optimal registration in the presence of uncertainties. There are broadly three steps (1) Approximate pose estimation (\APE), (2) Refinement for noise (\RN) and (3) Local dense refinement (\LDR). We use \TejasOne~\cite{izattglobally} or Go-ICP~\cite{Yang13} for \APE. We use only a subset of sensor points for \APE to get quick and approximate estimates. For \RN we use \TejasFive which is a mixed integer programming problem that we introduce in this work. Upon finding a refined solution, we further improve it using IMLP~\cite{Seth15} for \LDR. }
	\label{fg:flowchart}
\end{figure*}
Let us consider the problem of registering $N_s$ points obtained in the sensor frame to $N_m$ points in model frame. We restrict our current analysis to having a uniform isotropic noise represented by $\bC_i^s$, $i=1,\dots,N_s$ in the sensor points. We however, accommodate anisotropic noise in the model points, and assume they are drawn from a set of normal distributions each having covariance $\bC_j^m$, $j=1,\dots,N_m$. The uncertainty is lower in the direction of the local surface normal and higher along the tangential plane (refer to~\cite{segal2009generalized} for a detailed discussion on this.). Registration is posed as an optimization problem over the pose parameters, $\bR \in \SO,\bt \in \mathbb{R}^3$, and correspondence variables $ \corrFive \in \mathbb{R}^{N_s\times N_m}$ to minimize the sum of Mahalanobis distance between sensor points and corresponding model points,
\begin{align}
d_{i,j}^{2} & = (\bR \bss_i + \bt - \bm_j)^T \covariance_{ij}^{-1} (\bR \bss_i + \bt - \bm_j)\quad \text{where},\\
\covariance_{ij} &= (\covariance_j^m + \bR \covariance_i^s \bR^T ).
\end{align} 
The isotropic nature of the sensor noise makes $ \covariance_{ij}$ independent of $\bR$, since $\bR \covariance_i^s \bR^T=\covariance_i^s$ for $\covariance_i^s =\sigma_i\bI$. We perform Cholesky decomposition on $\covariance_{i,j}^{-1}$ to obtain lower and upper triangular matrix. $\chol_{i,j}$ and $\chol_{i,j}^T$ respectively, making it convenient to write the Mahalanobis distance in terms of the $l_1$ norm,
\begin{align}  
\label{mahalonbis dist}
 d_{i,j}^{Mah} & = \left|\chol_{i,j}^T   (\bR\bss_i + \bt - \bm_j)\right|_1.  
\end{align}
We prefer the distance in terms of $l_1$ norm as opposed to $l_2$ norm in order to retain a linear form of the equations. 
Note that this distance is different from the Euclidean distance used by ICP~\cite{ICP}, GoICP~\cite{Yang13} and Izatt~\etal~\cite{izattglobally}, \mbox{    	$d_{i,j}^{Eu}  = |\bR \bss_i + \bt - \bm_j|_1$  }.

The objective function to be minimized is obtained from the sum of Mahalanobis distances between all the sensor points and their corresponding model points. We assume that each sensor point can correspond to only one model point (this assumptions may not be true if there are outliers, which we deal with in Sec.~\ref{sc:outliers}), and impose this constraint by introducing a matching matrix $\corrFive \in \mathbb{R}^{\NsSampled \times N_m}$. Each element of this matrix, $\corrFive_{i,j}$, is a binary variable that takes value $1$ if $\bss_i$  corresponds to $\bm_j$. We formulate the optimization problem as shown below,
\begin{align}
\underset {\bR,\bt,\corrFive}{minimize}  &\sum_{i=1}^{\NsSampled} \sum _{j=1}^{N_m}  \left|\chol_{i,j}^T   (\bR\bss_i + \bt - \bm_j)\right|_1\corrFive_{i,j}\\
& \bR \in \SO, \nonumber \\
& \sum_{j=1}^{N_m} \corrFive_{i,j} = 1.\nonumber  
\end{align}
The constraints on $\bR$ are nonconvex and so we impose piecewise-convex relaxations to it by combining the approaches of \cite{dai2017global} and \cite{yu2018maximum}. More details can be obtained from Appendix~\ref{sc:RotConstraints}.

The bilinear terms in the objective function (for instance, the term $\corrFive_{i,j}\bt$) poses a challenge to most off-the-shelf MIP optimizers such as Gurobi~\cite{optimization2014inc}. In order to deal with this, we introduce additional variables $\beta_{i,j}\in\mathbb{R}^3$ and $\phi_{i,j}\in\mathbb{R}$ to \emph{`convexify'} the problem, 
\begin{align}
\label{eq:MIP1}
\underset {\bR,\bt,\corrFive}{minimize} & \sum_{i=1}^{\NsSampled} \sum _{j=1}^{N_m} \phi_{i,j}-(N_m-1)\times N_s \times \phimaxCorr, \\
\text{subject to}&\nonumber\\
&\text{Relaxed} \ R \in SO(3), \nonumber\\
\label{eq:betaconstraints}
&\beta_{i,j,k}\geq \pm (\chol_{i,j}^T(\bR\bss_i+\bt-\bm_j))_k,\\
&\phi_{i,j},\beta_{i,j,k}\geq 0,\nonumber\\
\label{eq:Hconstraints1}
&\phi_{i,j}\geq \sum_{k=1}^3 \beta_{i,j,k} - (1-\corrFive_{i,j})\mathds{M} ,\\
\label{eq:Hconstraints2}
&\phi_{i,j}\geq \phimaxCorr (1-\corrFive_{i,j}),\\
\label{eq:Hconstraints3}
&\sum_{j=1}^{N_m} \corrFive_{i,j} = 1.  .
\end{align}
where $\bv_k$ is the $k^{\text{th}}$ component of $\bv$, $\mathds{M}$ is an arbitrarily large number and $\phi_{max1}$ is a distance threshold for classifying points
as outliers~\footnote{In this work we choose $\mathds{M} = 10^{4}$, and $\phimaxCorr = 1000$.}. Note that we subtract the term $\phimaxCorr N_s (N_m-1)$ from $\sum_{ij} \phi_{i,j}$ in the objective function. This is done purely as a convenience of implementation. MIP optimizers often have a termination criteria which is the ratio of the difference between maximum bound  and minimum bound of the objective function value to maximum bound of objective function. If we do not subtract the $\phimaxCorr N_s (N_m-1)$ term, the solver might terminate before reaching the global minimum.

\subsection{Outlier detection}
\label{sc:outliers}  
    This formation can be easily extended to detect outliers by modifying Eq.~\ref{eq:Hconstraints3} as,  
         \begin{align}
            & \sum_{j=1}^{N_m} \corrFive_{i,j} + o_i = 1.\nonumber \\
            & \phi_{i,j} \geq o_i \phimaxOut  \nonumber \\
            & \phi_{i,j} \geq \sum_{k \in [1,3]} \beta_{i,j,k} - (1-\corrFive _{i,j})\mathds{M} - \mathds{M}o_i, \nonumber  
        \end{align}
        
    If the distance between sensor point and corresponding model point exceeds the threshold $\phimaxOut$ then these constraints assign the sensor point as an outlier assigning value $1$ to the variable $o_i$.

\subsection{Restricting correspondence search}
In most applications, there is a large number of model points and it may be wasteful to check for correspondences between each pair. Thus we can accelerate the optimization by either using heuristics (as in the case of~\cite{yu2018maximum}) or by first finding an approximate correspondence using the method of Izatt~\etal~\cite{izattglobally} or GoICP~\cite{Yang13}. We can then restrict the correspondence search to the models points that are close to the approximate correspondence obtained. We assume that we get an approximate range of possible corresponding points is given by $\bQ \in \mathbb{R}^{N_s \times \modelTolBandLength}$, where the $i^{\text{th}}$ row of the $\bQ$ contains indices of $\modelTolBandLength$ model points closest to $(\bR\bss_i+\bt)$. We replace $\corrFive$ with $\corrFivePntTwo \in \mathbb{R}^{\NsSampled \times \modelTolBandLength}$ in Eq.~\ref{eq:Hconstraints1}-\ref{eq:Hconstraints3}, and modify Eq.~\ref{eq:betaconstraints} as follows,
\begin{align*}
\beta_{i,j,k} \geq \pm \left( \chol_{i,\bQ_{i,j}}^T(\bR\bss_{i} + \bt - \bm_{\bQ_{i,j}})\right)_k
\end{align*}

\subsection{Multi-step Optimization}
Solving the MIP optimization as described in Eq.~\ref{eq:MIP1} is computationally expensive as we increase the total number of sensor points and model points. For example, when we ran \TejasFive for 100 sensor points and 100 model points, 64 GB RAM appeared to be insufficient to find the optimal solution. To overcome this problem, we break down the implementation into three parts -- (i) Approximate Pose Estimation~(\APE), Refinement for Noise~(\RN) and Local Dense Refinement~(\LDR). 
\par We require four points on a rigid body to uniquely define its 3D pose. This may not be necessarily true for every rigid body especially when the object is symmetric or the points obtained have noise in them. Through empirical observations, Srivatsan~\etal~\cite{srivatsan2017sparse} have observed that about 20 sensor points are sufficient for most shapes, in order to obtain a reasonable registration estimate. In $\APE$, we randomly select a small subset of sensor points to register with the model points using $\TejasOne$ along with ICP to provide heuristics. Since $\TejasOne$ optimizes an objective function with Euclidean distance, without accounting for the uncertainty parameters, it has fewer number of optimization variables and quickly provides an approximate pose within a few degrees of misalignment. We use this pose and find an approximate band of $\modelTolBandLength$ model points for correspondence calculation for each sensor point. 
\par In \RN, we restrict our correspondence search using the solution provided by \APE and optimize using $\TejasFive$. Since we are using only a subset of sensor points, it is possible that the solution obtained from \TejasFive may not be globally optimal. We select a number of candidate solutions provided by the optimizer ranked by the value of the objective function at those solutions. We then perform a local refinement on all these candidate solutions and select the pose having least objective function value.   While performing the \LDR we use all the sensor points.

%% file: results.tex
\begin{figure*}[h!]
	\centering
	\includegraphics [width=0.8\textwidth]
	{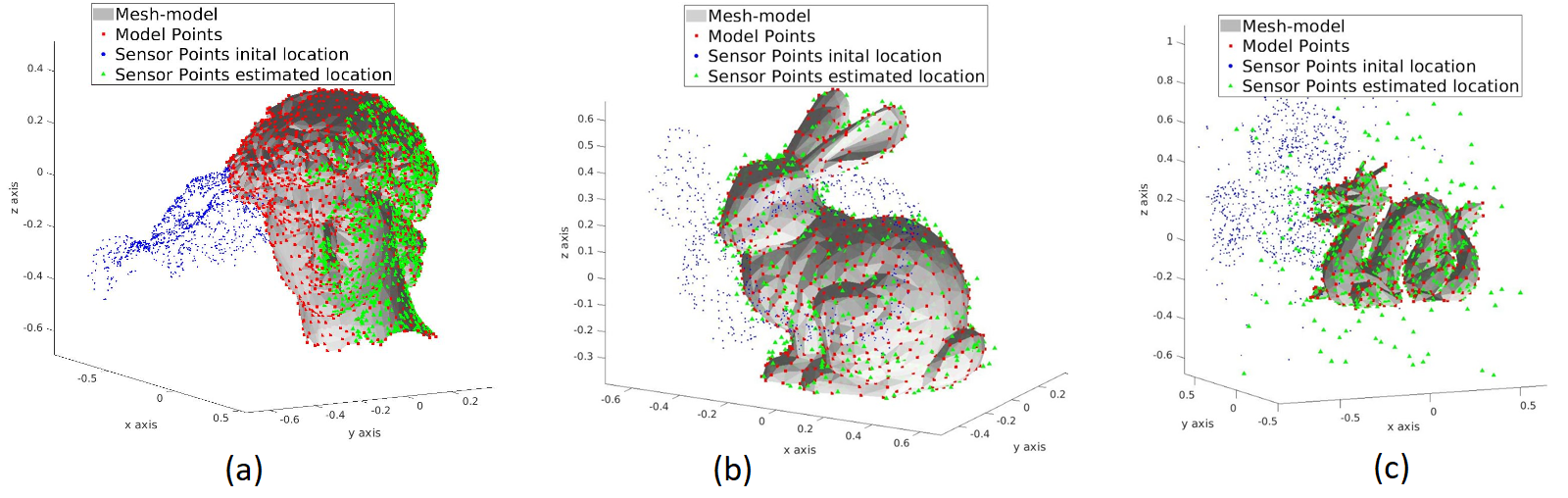}
	\caption{ The mesh model is shown in grey, model points are shown in red, initial location of sensor points is shown in blue and estimated locations of registered sensor points is shown in green. (a) Registration with partial overlap for the head of David~\cite{levoy2000digital}, (b) registration in the presence of noise for Stanford Bunny~\cite{turk2005stanford}, (c) registration in the presence of outliers for Dragon~\cite{turk2005stanford}. In all the three scenarios, our approach finds globally optimal solutions.    }
	\label{fg:model}
\end{figure*}

\section{Results}
\label{sc:results}
We implemented our approach on Intel - Core i9-7940X 3.1GHz 14-Core Processor in python 2.7 with Gurobi 7.5.2 for mixed integer branch and bound optimization. While generating synthetic data, for model points, we sampled 1 to 10 points from each face of the mesh model. Around 1000 Sensor points were sampled independent of the model points and an isotropic Gaussian noise was added to each sensor point. These sensor points were then transformed with a random, but known ground truth transformation to check accuracy of the solution given by our method.   
 \input{tableband.tex}
\subsection{Robustness to varying correspondence bands}
We sub-sampled 20 sensor points for the \APE using \TejasOne. We set the maximum run time for \TejasOne to be 300 seconds (though the solver converged to a solution in much less than this time in most cases) which provides pose estimates with a rotation error of approximately $3^\circ$ from the ground truth. It is common practice to use heuristics to hasten the search for optimal solutions in MIP implementations~\cite{izattglobally}. In our implementation, every time the solver found a feasible solution, we used ICP (with approx 300 sensor and 500 model points) as heuristics, and provided the solution back to \TejasFive for efficient branch and bound search. In this section, we present results for registration of a partial point cloud to a model of the head of David~\cite{levoy2000digital} as shown in Fig.~\ref{fg:model}(a). We tested our method for band sizes of 5, 20, 30 and 50. Table~\ref{tb:band} shows the error in the rotation (in degrees), error in translation, target registration error (TRE)~\footnote{TRE is the average Euclidean distance between the registered and known ground-truth positions of the sensor points.}, and value of the objective function in Eq.~\ref{eq:MIP1}. These results are compared with ICP, GICP, IMLP, \TejasOne and GoICP~\footnote{GoICP was implemented in C++ while other methods were implemented in python.}. As we increase the band size, the chances of \TejasFive to get stuck in a local valley are high. But even then, the \LDR helps to get closer to the ground truth, when compared to the other methods.  It is worth noting that the misalignment for this experiment was very high that ICP and IMLP failed to estimate the registration altogether. The average time taken by ICP is 0.1s, IMLP is 10.2s, GoICP is 1.5min, \TejasFive is 5min. The IMLP used in \LDR only takes 1.2s on an average. Instead of choosing a band of correspondences of fixed size, one could also restrict the correspondences based on a distance threshold. 
Note that in Table~\ref{tb:band}, \TejasFive can also be interpreted as \APE+\RN. As illustrated in Fig.~\ref{fg:optimal}, it is clear that the multistep process helps produce the most accurate solution to this problem. As interesting observation we make from the last column of Table~\ref{tb:band} is that, the value of the objective function ( see Eq.~\ref{eq:MIP1}) is lowest for our approach, even lower than the value at the ground truth registration. Infact, the value of objective function is most similar to the ground truth for GoICP. One must be cautious about this observation because, Izatt~\etal~\cite{izattglobally} and Yu and Ju~\cite{yu2018maximum} have observed scenarios where GoICP produced poor results compared to \TejasOne.

\input{tablenoise.tex}

\subsection{Robustness to varying levels of noise}
We tested our method for noise levels varying from $\sigma =5 \times 10^{-5} $ to $ 4\times 10^{-2}$ in a model restricted to fit in a  $1\times1\times1$ unit box. We select a band of around 20 model points neighboring the corresponding model point obtained from the APE solution. For this experiment we consider a Stanford bunny~\cite{turk2005stanford} (Fig.~\ref{fg:model}(b) shows the result for $\sigma=10^{-2}$). From Table~\ref{tb:noise}, we observe that the accuracy of all the methods decrease as the noise is increased. Among all the methods, \TejasFive consistently provides good estimates, which are then refined by \LDR taking it closer to the global optima. We also observe that the solution obtained by \LDR after \APE is not as accurate as \APE followed by \RN and then \LDR. Another point to note is that the average time taken for GoICP was much higher compared to the previous experiments with no noise. GoICP took an average of 15 min to find a solution and sometimes required tuning of tolerance parameters to produce any solution in reasonable time.

\begin{figure*}[htbp]
	\centering
	\includegraphics [width=0.8\textwidth]
	{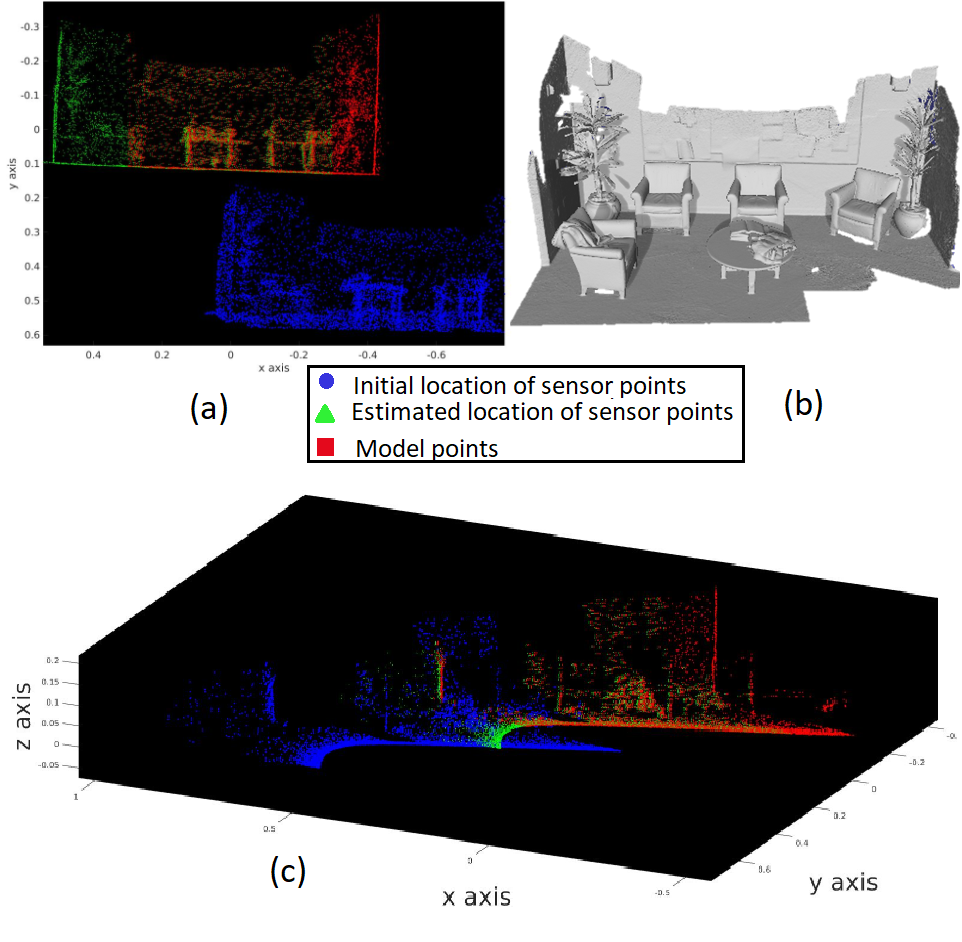}
	\caption{ Model points are shown in red, initial location of sensor points is shown in blue and estimated locations of registered sensor points is shown in green. (a) Registration result on realworld RGBD dataset of a lounge. We only use the point cloud locations and neglect the RGB information. Note that the green and red points are noisy and have partial overlap, yet our approach is able to register them accurately. (b) Rendered view of the lounge dataset from~\cite{zhou2013dense}. (c) Registration of point clouds obtained from a velodyne lidar from the Oakland dataset~\cite{munoz2009contextual}.  }
	\label{fg:lidarindoor}
\end{figure*}

\subsection{Robustness to partial overlap and outliers}
To check for the robustness of our approach to outliers, we tested it on synthetic data of a dragon~\cite{turk2005stanford} as shown in Fig.~\ref{fg:model}(c). We used a variant of ICP called Trim-ICP~\cite{chetverikov2005robust} to deal with outliers, when used as a heuristic in \APE. Keeping the initial misalignment constant, we added different percentages of outliers and tested the robustness of our approach. Table~\ref{tb:outliers} shows the errors for various percentage of outlier data. We observe that our approach is robust to presence of outliers, which is critical when used with real world data.

\input{tableoutliers.tex}

We also tested our approach on two real world data sets. The first is an example of 3D reconstruction from RGBD point cloud. For this experiment we considered RGBD scans  of a lounge~\cite{zhou2013dense} (See Fig.~\ref{fg:lidarindoor} (b)
). We ignored the color information and only used the point cloud information. Fig.~\ref{fg:lidarindoor}(a) shows the initial position of point clouds obtained from two different views in red and blue colors. Based on a prior knowledge of the sensor used, we use an approximate sensor uncertainty in the blue points. For the red points, we use PCA analysis similar to the work of Esterpar~\etal~\cite{estepar2004robust}, to find the covariance along local surface normal and the tangential plane. The estimated error in the rotation is $2.09^\circ$ and translation is $0.28$cm. The estimated location of the points, shown in green aligns well with the red points, despite presence of only a partial overlap, as shown in Fig.~\ref{fg:lidarindoor}(a). 

We repeat this analysis for another experiment involving realworld measurement obtained from a Velodyne lidar sensor.
We use scans from the Oakland dataset for this experiment~\cite{munoz2009contextual}. The estimated error in the rotation is $0.38^\circ$ and translation is $0.7$cm. Just to show the versatility of our framework, we use GoICP for \APE in this example. Once again we observe that our approach registers the red and blue point clouds and produces an accurate alignment as shown by the green points in Fig.~\ref{fg:lidarindoor}(c).

%% file: tableband.tex
\setlength{\tabcolsep}{0.2mm}
\renewcommand{\arraystretch}{0.5}
\begin{table}[htbp!]
\caption{Results for varying Correspondence band sizes}
\label{tb:band}
\centering
\begin{tabular}{l c c c c r }
\toprule
{}&Band size&Rot. & Trans.  & TRE &Obj val  \\
{}& & error (deg)& error  & & \\
 \midrule
Ground truth& {}&0&0&0&449.60\\
ICP&{} &141.09&1.14&0.49&20564.99\\
IMLP&{} &112.65&0.944&0.46&2623.41\\
GoICP&{} &0.33&$2\times 10^{-3}$&$1.2\times 10^{-3}$&408.41\\
\TejasOne &{}&2.93&0.02&0.01&2696.55\\
\midrule 
\TejasFive&  \multirow{2}{*}{5}  &0.10&$2\times 10^{-4}$&$8.2\times10^{-5}$&173.72\\
\APE+\RN+\LDR &{}&0.10&$2 \times 10^{-5}$&$1.55\times 10^{-5}$&164.93\\
\midrule 
\TejasFive&\multirow{2}{*}{20}  &1.26&0.01&$7\times 10^{-3}$&719.81\\
\APE+\RN+\LDR&{} &0.10&$2\times10^{-5}$&$1.5\times10^{-5}$&164.93\\

\midrule 
\TejasFive&\multirow{2}{*}{30}  &1.22&0.01&$7\times 10^{-3}$&720.02\\
\APE+\RN+\LDR&{} &0.10&$2\times10^{-5}$&$1.5\times10^{-5}$&164.93\\
\midrule 
\TejasFive&\multirow{2}{*}{50}  &1.28&0.01&$7\times 10^{-3}$&731.66\\
\APE+\RN+\LDR&{} &0.10&$2\times10^{-5}$&$1.5\times10^{-5}$&164.93\\
\bottomrule
\end{tabular}
\end{table}


%% file: tablenoise.tex
\setlength{\tabcolsep}{1mm}
\renewcommand{\arraystretch}{0.5}
\begin{table}[htbp!]
\caption{Results for varying noise}
\label{tb:noise}
\centering
\begin{tabular}{l c c c c }
\toprule
{}&Rot. error ($^\circ$)& Trans. error  & TRE &Obj val\\ \midrule
\multicolumn{5}{c}{$\sigma=5\times10^{-5}$}\\
Ground truth &0&0&0&28394.42\\
ICP &127.16&1.31&0.56&331829.18\\
IMLP &129.36&1.25&0.60&1158433.99\\
GoICP &0.30&$4.2\times10^{-3}$&$1.8\times10^{-3}$&28778.12\\
\TejasOne &1.76&0.02&0.012&42562.80\\
\TejasFive &0.23&$1\times10^{-3}$&$9.9\times10^{-4}$&28616.78\\
\APE+\RN+\LDR &0.23&$2.8\times10^{-3}$&$9.8\times10^{-4}$&28245.36\\
\midrule
\multicolumn{5}{c}{$\sigma=10^{-4}$}\\
Ground truth &0&0&0&13988.28\\
ICP &159.51&	0.43&0.41&104004.00\\
IMLP &143.74&0.37&0.51&370540.86\\
GoICP &0.32&$2\times10^{-3}$&$2\times10^{-3}$&14131.38\\
\TejasOne &3.72&0.02&0.02&30430.84\\
\TejasFive &0.27&$2.2\times10^{-3}$&$2.3\time10^{-3}$&14592.16\\
\APE+\RN+\LDR &0.22&$1.5\times10^{-3}$&$1.5\times10^{-3}$&13983.39\\
\midrule
\multicolumn{5}{c}{$\sigma=5\times10^{-3}$}\\
Ground truth &0&0&0&370.39\\
ICP &77.35&0.62&0.28&2275.62\\
IMLP &61.36&0.51&0.24&1981.30\\
GoICP &0.67&$5\times10^-3$&$4\times10^-3$&385.69\\
\TejasOne &2.94&0.02&0.01&553.57\\
\TejasFive &0.29&$3.5\times10^{-3}$&$2\times10^{-3}$&372.78\\
\APE+\RN+\LDR &0.14&$2\times10^{-3}$&$1\times10^{-3}$&368.35\\
\midrule
\multicolumn{5}{c}{$\sigma= 10^{-2}$}\\
Ground truth &0&0&0&240.77\\
ICP &176.09&0.83&0.62&1229.01\\
IMLP &172.39&0.84&0.621120.02\\
GoICP &1.08&$9\times10^{-3}$&$5.4\times10^{-3}$&242.56\\
\TejasOne &1.78&0.01&$7\times10^{-3}$&260.43\\
\TejasFive &1.18&$9\times10^{-3}$&$6\times10^{-3}$&241.95\\
\APE+\RN+\LDR &0.62&$4\times10^{-3}$&$5\times10^{-3}$&235.48\\
\midrule
\multicolumn{5}{c}{$\sigma=4\times 10^{-2}$}\\
Ground truth &0&0&0&166.53\\
ICP &55.00&0.35&0.31&445.86\\
IMLP &54.69&0.35&0.30&381.18\\
GoICP &3.70&0.04&0.02&158.62\\
\TejasOne &5.09&0.04&0.02&170.65\\
\TejasFive &5.30&0.04&0.02&158.97\\
\APE+\RN+\LDR &2.33&0.02&0.01&157.26\\
\bottomrule
\end{tabular}
\end{table}

%% file: tableoutliers.tex
\setlength{\tabcolsep}{1mm}
\renewcommand{\arraystretch}{0.5}
\begin{table}[htbp!]
\caption{Results for varying levels of outliers}
\label{tb:outliers}
\centering
\begin{tabular}{l c c c c }
\toprule
Outlier ($\%$)&Rot. error ($^\circ$)& Trans. error  & TRE \\
 \midrule
10 &0.80&$4\times10^{-3}$&$5\times10^{-3}$\\
20 &0.95&$5\times10^{-3}$&$6\times10^{-3}$\\
40 &0.49&$4\times10^{-3}$&$3\times10^{-3}$\\
\bottomrule
\end{tabular}
\end{table}

%% file: conclusion.tex
\section{Discussions and Future Work}
\label{sc:conclusion}
In this work, we presented a mixed integer programming (MIP) based approach for globally optimal registration, while considering uncertainty in the point measurements. We observe that our approach is effective at finding optimal solutions are various level of noise and outliers as well as effectively deal with partial overlap in the data. Our implementation involves multi-step optimization, which allows for fast computation without compromising on the quality of the final result. We believe that in applications where registration accuracy is of great importance and real-time performance is not critical, our approach can be very effective in providing a benchmark for finding optimal solutions. 

The implementation can be made faster by incorporating application specific local heuristic solvers. We demonstrated improved performance using a few popular local approaches, however, our framework is flexible enough to use any other local heuristics methods. As exciting future direction that we are currently pursuing involves using data-driven techniques to provide fast heuristics for the MIP solvers without the requirement for correspondence calculations.

 Currently, we use an off-the-shelf MIP optimizer which is a generalized solver and not specifically optimized for our problem. Since we know the structure of our problem, we believe that the branch and bound search strategy could be modified to effectively find solutions for our problem. 
In the future we plan to relax the isotropic uncertainty assumption in the sensor points and extend the approach to consider anisotropic uncertainty. We also plan to incorporate surface normal and curvature information as well as introduce a regularization term to perform deformable registration. 

%% file: appendix.tex
\subsection{Rotation matrix constraints}
\label{sc:RotConstraints}
Consider the rotation matrix \mbox{$\bR = [\textbf{u}_1, \textbf{u}_2, \textbf{u}_3]^T \in SO(3)$}, where $\bu_i$ are orthogonal unit vectors. There exist the following constraints on $\bR$, $\bR^T\bR=\bI$ and $\det (\bR)=1$. Since these constraints add nonconvexity to the problem, we approximate the constraint on each element of $\bR$ with piecewise-convex approximations~\cite{}.
\paragraph{Orthogonality constraints}
In order to approximate the orthogonality constraints $\bR^T\bR=\bI$, we divide the range $[-1,1]$ in $n$ intervals with $k^{\text{th}}$ interval being $[q_k,q_{k+1}]$. We introduce auxiliary variables $ \textbf{w} \in \mathbb{R}^{3x3}$ and $\lambda^{i,j} \in  \mathbb{R}^{n+1}$ such that,
\begin{align*}
&\begin{bmatrix}
\textbf{u}_i(j) \\
\text{\textbf{w}}_i(j) \\
\end{bmatrix}
= \sum_{k=0}^n \lambda_k^{i,j}
\begin{bmatrix}
q_k \\
q_k^2 \\
\end{bmatrix}, \lambda^{i,j} \text{in sos2}\\
&\textbf{u}_i^T\textbf{u}_i \leq 1  \\
&\text{\textbf{w}}_i(1)+\text{\textbf{w}}_i(2)+\text{\textbf{w}}_i(3) \geq 1,  \\
&|\textbf{u}_i \pm \textbf{u}_j|_2^2 \leq 2  \\
&|\textbf{u}_1 \pm \textbf{u}_2 \pm \textbf{u}_3|_2^2 \leq 3. 
\end{align*}
For a discussion on sos2 constraints, refer to~\cite{dai2017global}. In this work we choose the number of partitions for the sos2 constraints to be 50. Increasing the number of partitions improves the result but also increases the computation time. 
\paragraph{Bi-linear terms }
In order to impose the constraint $\det (\bR)=1$, which can also be written as $\textbf{u}_i \times \textbf{u}_j = \textbf{u}_k$, we use McCormick constraints~\cite{mccormick1976computability},
This constraint involves bi-linear terms (for e.g. $\mathbf{R}_{3,1} = \mathbf{R}_{1,2}\mathbf{R}_{2,3} + \mathbf{R}_{2,2}\mathbf{R}_{1,3}$) and the non-linearity makes the problem difficult to solve. We apply the relaxation of every such bi-linear term $xy$ where $x$ and $y$ are two continuous variables such that $x\in [x_{lb},x_{ub}]$ and $y\in [y_{lb},y_{ub}]$. For rotation matrix components, these upper bounds and lower bounds are $1$ and $-1$ respectively. We approximate each bi-linear term $xy$ by a new scalar variable $v$. We apply McCormick relaxation and find out over (concave) and under (convex) envelop function for the bi-linear term. This gives $v$ as a close linear approximation of the bi-linear term $xy$. 
\begin{align*}
concave(xy) &= min (-y + x +1 ,  y-x+1) \\
convex(xy) &= max(-y - x +1,  y+ x-1) \\
v &\geq -y-x-1,  \\
v &\geq  y+ x-1, \\
v &\geq  y-x+ 1, \\
v &\geq -y+ x+1,\\ 
 convex(xy) &\leq v \leq concave(xy).
\end{align*}

We use the following convention,
\begin{align*}
v_1&= \bR_{1,3}\bR_{2,2}, \\
v_2&= \bR_{1,2}\bR_{2,3},  \\
v_3&= \bR_{2,1}\bR_{1,3},  \\
v_4&= \bR_{1,1}\bR_{2,3},  \\
v_5&= \bR_{2,1}\bR_{1,2},  \\
v_6&= \bR_{1,1}\bR_{2,2}.
\end{align*}
Using these approximations for each bi-linear term, each element of the rotation matrix can be approximated. For example, since $\bR \in SO(3)$, $\bR_{3,1} = \bR_{1,2}\bR_{2,3} - \bR_{2,2}\bR_{1,3}$ (cross product of first and second column equals third column), $\bR_{3,1} =  (v_2 - v_1)$. The other elements can be similarly obtained.